\documentclass[11pt,a4paper]{article}
\usepackage[hyperref]{aacl-ijcnlp2020}
\usepackage{times}
\usepackage{latexsym}

\usepackage{microtype}
\usepackage{graphicx}
\usepackage{xcolor}
\usepackage{multirow}
\usepackage{graphicx}
\usepackage{amsmath}
\usepackage{amssymb}
\usepackage{amsthm}
\usepackage{comment}
\usepackage{booktabs}

\aclfinalcopy 

\title{Generating Commonsense Explanation by Extracting Bridge Concepts from Reasoning Paths}

\author{Haozhe Ji$^1$, Pei Ke$^1$, Shaohan Huang$^2$, Furu Wei$^2$, Minlie Huang$^1$\thanks{\quad Corresponding author}  \\
$^1$Department of Computer Science and Technology,
Institute for Artificial Intelligence, \\
State Key Lab of Intelligent Technology and Systems, \\ 
Beijing National Research Center for Information Science and Technology, \\ 
Tsinghua University, Beijing 100084, China \\
$^2$Microsoft Research \\
  {\tt\small \{jhz20, kp17\}@mails.tsinghua.edu.cn, \{shaohanh, fuwei\}@microsoft.com, } \\
  {\tt\small aihuang@tsinghua.edu.cn } \\}

\date{}

\begin{document}
\maketitle
\begin{abstract}
Commonsense explanation generation aims to empower the machine's sense-making capability by generating plausible explanations to statements against commonsense. While this task is easy to human, the machine still struggles to generate reasonable and informative explanations. In this work, we propose a method that first extracts the underlying concepts which are served as \textit{bridges} in the reasoning chain and then integrates these concepts to generate the final explanation. To facilitate the reasoning process, we utilize external commonsense knowledge to build the connection between a statement and the bridge concepts by extracting and pruning multi-hop paths to build a subgraph. We design a bridge concept extraction model that first scores the triples, routes the paths in the subgraph, and further selects bridge concepts with weak supervision at both the triple level and the concept level. We conduct experiments on the commonsense explanation generation task and our model outperforms the state-of-the-art baselines in both automatic and human evaluation. 
\end{abstract}

\section{Introduction}

Machine commonsense reasoning has been widely acknowledged as a crucial component of artificial intelligence and a considerable amount of work has been dedicated to evaluate this ability from various aspects in natural language processing~\cite{Levesque2011TheWS,Talmor2018CommonsenseQAAQ,Sap2019SocialIC}. A large proportion of existing tasks frame commonsense reasoning as multi-choice reading comprehension problems, which lack direct assessment to machine commonsense~\cite{Wang2019DoesIM} and impede its practicability to realistic scenarios~\cite{Lin2019CommonGenAC}. Recently, \citet{Wang2019DoesIM} proposed a commonsense explanation generation challenge that directly tests machine's sense-making capability via commonsense reasoning. In this paper, we focus on the challenging explanation generation task where the goal is to generate a sentence to explain the reasons why the input statement is against commonsense, as shown in Figure \ref{fig:bridge}.

Generating a reasonable explanation for a statement faces two main challenges: 1) \textbf{Trivial and uninformative explanations}. 
As this task can be formulated as a sequence-to-sequence generation task, existing neural language generation models tend to generate trivial and uninformative explanations. For example, one of the existing neural models generates an explanation ``\textit{The school wasn't open for summer}'' to the statement in Figure \ref{fig:bridge}. Although it is sometimes reasonable, simple modification of the statement to the negation form with no additional information cannot explain the reasons why the statement conflicts with commonsense. 2) 
\textbf{Noisy commonsense knowledge grounding}. It's still challenging for most existing language generation models to generate explanations that are faithful to commonsense \cite{Lin2019CommonGenAC}. Thus, explicitly incorporating external knowledge sources is necessary for this task. Since the nature of the explanation generation task involves using underlying commonsense knowledge to explain, locating useful commonsense knowledge from large-scale knowledge graph is not trivial and generally requires multi-hop reasoning.

\begin{figure}[t]
    \centering
    \small
    \begin{tabular}{l}
    \hline
    \textbf{Statement:} The \textcolor{green!70!black}{\textit{school}} was open for \textcolor{green!70!black}{\textit{summer}}.\\
    \textbf{Explanation:} \textcolor{green!70!black}{\textit{Summertime}} is typically \textcolor{blue}{\textit{vacation}} time \\
    for \textcolor{green!70!black}{\textit{school}}.\\
    \hline  
    \end{tabular}
    \caption{Generating a reasonable and informative explanation involves generating \textit{bridge concepts} like \textcolor{blue}{\textit{vacation}} by identifying the relation to the \textit{source concepts}, i.e. \textcolor{green!70!black}{\textit{school}} and \textcolor{green!70!black}{\textit{summer}} in the statement. 
    }
    \label{fig:bridge}
\end{figure}

To address the above challenges, we propose a two-stage generation framework that first extracts the critical concepts served as \textit{bridges} between the statement and the explanation from an external commonsense knowledge graph, and then generates plausible explanations with these concepts. We first retrieve multi-hop reasoning paths from ConceptNet~\cite{speer17conceptnet} and heuristically prune the paths to maintain the coverage to plausible concepts while keeping the scale of the subgraph tractable. Before the extraction stage, we initialize the representation of each node on the subgraph by fusing both the contextual and graph information. Then, we design a bridge concept extraction model that scores triples, propagates the probabilities along multi-hop paths to the connected concepts and further extracts plausible concepts. In the second stage, we use a pre-trained language model~\cite{Radford2019LanguageMA} to generate the explanation by integrating both the statement and the extracted concept representations. 
Experimental results show that our framework outperforms knowledge-aware text generation baselines and GPT-2~\cite{Radford2019LanguageMA} in both automatic and human evaluation. Particularly, our model generates explanations with more informative content and provides reasoning paths on the knowledge graph for concept extraction.

To summarize, our contributions are two-fold:
\begin{itemize}
    \item We analyze the under-explored commonsense explanation generation task and investigate the challenges in incorporating external knowledge graph to aid the generation problem. To the best of our knowledge, this is the first work on generating explanations for counter-commonsense statements.
    
    \item We propose a two-stage generation method that first extracts the bridge concepts 
    from reasoning paths and then generates the explanation based on these concepts. Our model outperforms state-of-the-art baselines on the commonsense explanation generation task in both automatic and human evaluation.
\end{itemize}

\begin{figure*}[t]
    \centering
    \includegraphics[width=2.0\columnwidth]{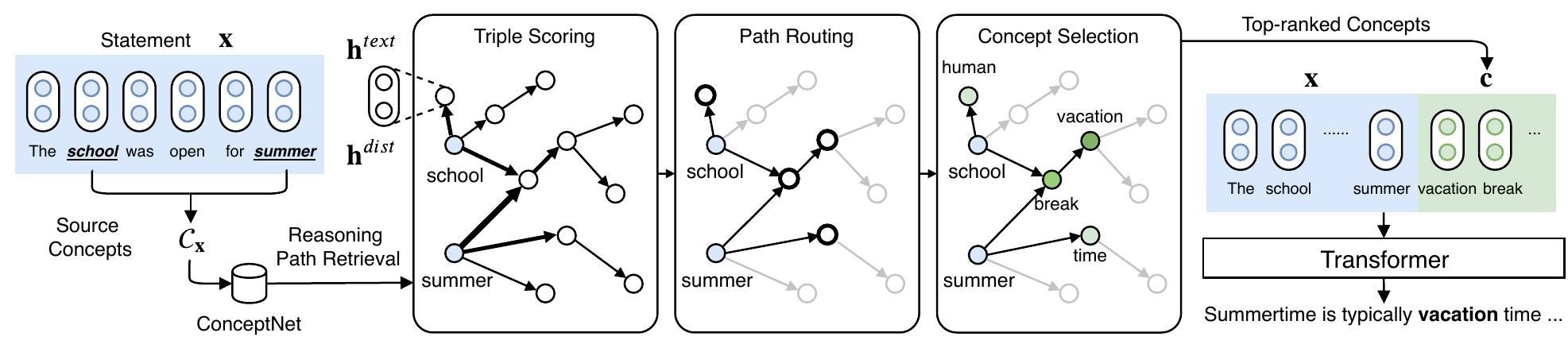}
    \caption{The inference process of our model. In the reasoning path retrieval stage (\S{\ref{sec:path-grounding}}), a subgraph is firstly retrieved from the ConceptNet given the source concepts ($\mathcal{C}_\mathbf{x}$), where each node representation is fused with both textual and graph-aware representations (\S{\ref{sec:repr}}). Then the model scores each triple on the subgraph, routes the path by propagating the probabilities along paths to the connected nodes, and selects concepts from activated nodes (\S{\ref{sec:concept-extraction}}). Finally, the model generates the explanation by integrating the token embeddings of both the statement and the top-ranked concepts (\S{\ref{sec:generation}}).
    }
    \label{fig:model}
\end{figure*}

\section{Related Work}

\subsection{Machine Commonsense Reasoning}
Previous work on machine commonsense reasoning mainly focuses on the tasks of inference~\cite{Levesque2011TheWS}, question answering~\cite{Talmor2018CommonsenseQAAQ,Sap2019SocialIC} and knowledge base completion~\cite{Bosselut2019COMETCT}. While the ultimate goals of these tasks are different from ours, we argue that performing explicit commonsense reasoning is also critical to generation. A line of work~\cite{Bauer2018CommonsenseFG,Lin2019KagNetKG} resorts to structured commonsense knowledge and builds graph-aware representations along with the contextualized word embeddings to tackle the commonsense question answering problem. 
In our work, we focus on reasoning over structured knowledge to explicitly infer discrete bridge concepts that are further used for text generation. Another line of work~\cite{Rajani2019ExplainYL,Khot2019WhatsMA} identifies the knowledge gap critical for the complete reasoning chain and fills the gap by writing general explanation or acquiring fine-grained annotations with human effort. While sharing a similar motivation, our method differs from theirs in the sense that we acquire distant supervisions for the bridge concepts 
to extract reasoning paths and generate plausible explanations without the need of additional human annotation.

\subsection{Knowledge-grounded Text Generation}
Existing work that utilizes structured knowledge graphs to generate texts mainly lies in conversation generation~\cite{zhou2018commonsense,dykgchat,opendialkg}, story generation~\cite{guan2019story} and language modeling~\cite{ahn2016neural,logan-etal-2019-baracks,hayashi2019latent}. \citet{zhou2018commonsense} and \citet{guan2019story} propose to use graph attention that incorporates the information of neighbouring concepts into context representations to help generate the target sentence.
Since one-hop graphs from concepts in the statement have low coverage to the concepts in the target explanation, this method is not directly applicable in our task.
Another direction that utilizes more complex graph is to model multi-hop reasoning by performing random walk~\cite{opendialkg} on the knowledge graph or simulating a Markov process on the pre-extracted knowledge paths~\cite{dykgchat}. While in our task, we don't have access to a parallel grounded knowledge source nor the bridge concepts, which makes the problem even more challenging.

\section{Methodology}

\subsection{Task Definition}

The commonsense explanation generation task is defined as generating an explanation given a statement against commonsense. Let $\mathbf{x}=x_1\cdots x_N$ be the input statement with $N$ words and $\mathbf{y}=y_1\cdots y_M$ be the explanation with $M$ words. A simple sequence-to-sequence formulation which learns a mapping from $\mathbf{x}$ to $\mathbf{y}$ can be adopted in this task:
\begin{equation}\label{equ:x2y}
    P(\mathbf{y}|\mathbf{x}) = \prod_{t=1}^{M}P(\mathbf{y}_{t}|\mathbf{y}_{<t},\mathbf{x}).
\end{equation}

\subsection{Model Overview}

Formally, our model generates the explanation by firstly extracting the critical bridge concepts $\mathbf{c}$ on a retrieved knowledge graph $G_\mathbf{x}$ given the statement $\mathbf{x}$ and then integrating the bridge concepts and the statement to generate a proper explanation $\mathbf{y}$, which can be formulated as follows:
\begin{equation}
    P(\mathbf{y}, \mathbf{c}|\mathbf{x}) =  P(\mathbf{c} | \mathbf{x}) P(\mathbf{y}|\mathbf{x}, \mathbf{c})
\end{equation}
where the bridge concepts $\mathbf{c}$ are defined as the unique concepts delivered in the explanation but not mentioned in the statement. Figure \ref{fig:model} presents the overview of our model framework. Firstly, we retrieve multi-hop reasoning paths from the ConceptNet based on the statement, and heuristically prune the noisy connections to obtain a subgraph for further concept extraction (\S{\ref{sec:path-grounding}}). To score the paths and concepts, we obtain the fused concept representation for each node on the subgraph by considering both the contextual and graph information (\S{\ref{sec:repr}}).
Secondly, we design a path routing algorithm to propagate the triple probabilities along multi-hop paths to the connected concepts and further extract plausible concepts (\S{\ref{sec:concept-extraction}}). 
Finally, our model generates the explanation by integrating the statement representation and the selected concept representation as inputs (\S\ref{sec:generation}).

\subsection{Reasoning Path Retrieval} \label{sec:path-grounding}
In this section, we demonstrate how we 
retrieve and prune the reasoning paths to form a subgraph. We also acquire distant supervision for uncovering the bridge concepts in the subgraph to supervise the concept extraction in the next stage.

Given an external commonsense knowledge graph $G=(V, E)$,
for each statement $\mathbf{x}$, we extract source concepts $\mathcal{C}_\mathbf{x} = \{c^i_{\mathbf{x}}\}$ from $\mathbf{x}$ by aligning the surface texts in $\mathbf{x}$ to the concepts in $V$. We also use the stem form of the surface texts to enable soft alignment and filter out stop words. At the training phase, we extract the target concepts $\mathcal{C}_\mathbf{y} = \{c^j_{\mathbf{y}}\}$ from the explanation $\mathbf{y}$ with a similar procedure.

Starting with the source concepts, we then retrieve reasoning paths from the knowledge graph to form a subgraph that has relatively \textbf{high coverage} to the bridge concepts with a \textbf{tractable scale}. 

\begin{figure}[t]
    \centering
    \includegraphics[width=\columnwidth]{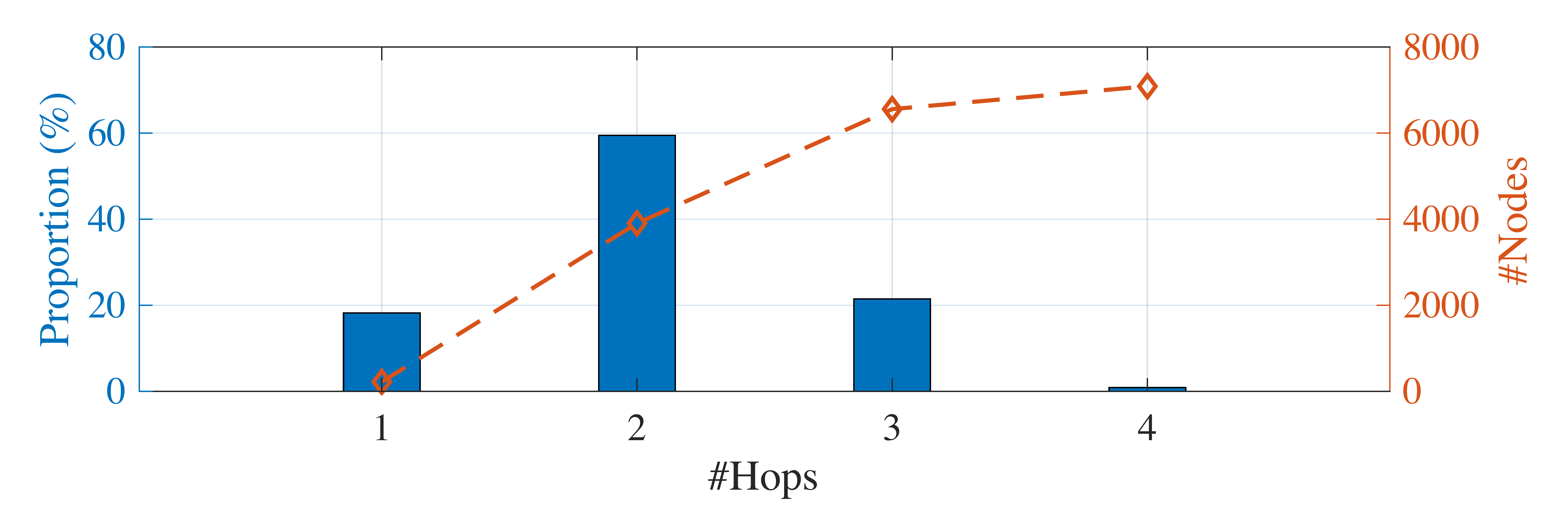}
    \caption{The left axis presents the distribution of the minimum required number of hops to reach the concepts in the explanation set $\mathcal{C}_{\mathbf{y}} - \mathcal{C}_{\mathbf{x}}$ from the source concepts in $\mathcal{C}_{\mathbf{x}}$. The right axis shows the number of nodes in the subgraph with different number of hops.}
    \label{fig:hops} 
\end{figure}

We first examine the minimum length of paths that connect source concepts $\mathcal{C}_{\mathbf{x}}$ with each concept in the explanation set $\mathcal{C}_{\mathbf{y}} - \mathcal{C}_{\mathbf{x}}$.  
As shown in Figure \ref{fig:hops}, over $80\%$ of the examples require two or three hops of connection from the source concepts to the concepts that are merely mentioned in the explanation, which indicates the necessity for multi-hop reasoning. 

We then count the number of concepts covered by subgraphs with different numbers of hops starting from the source concepts (We only consider concepts in the training data). As Figure \ref{fig:hops} shows, the average number of nodes covered by 3-hop subgraph exceeds 6,000, indicating the need of path pruning to keep the scale tractable.

Therefore, we design a heuristic algorithm to retrieve a subgraph $G_\mathbf{x} =\{V_\mathbf{x}, E_\mathbf{x}\}$ from the ConceptNet by expanding the source concepts with 3 hops to cover most bridge concepts. 
To keep the scale of the subgraph tractable, at each iterating step, we enlarge $V_\mathbf{x}$ with $B$ neighbour concepts most commonly visited by concepts in $V_\mathbf{x}$. Intuitively, the salient bridge concepts should be in a reasonable distance from the source concepts on the graph to maintain the semantic relation and should be commonly visited nodes that support the information flow on the graph. 

We distantly label the bridge concepts as the unique concepts in the explanation that could be covered by the subgraph:
\begin{equation}
    \mathcal{B}_{\mathbf{x}\rightarrow\mathbf{y}} = \{c|c\in \mathcal{C}_\mathbf{y} - \mathcal{C}_\mathbf{x}, c \in V_\mathbf{x}\}
\end{equation}

\subsection{Fused Concept Representation}\label{sec:repr}

We initialize each node on the subgraph with a fused concept representation $\mathbf{h}_c$ by considering both the contextual feature of the concept and the graph-aware information. We first obtain the contextualized statement representation $\mathbf{H}_\mathbf{x}\in \mathbb{R}^{N\times d_1}$ using a multi-layer bi-directional Transformer encoder~\cite{Vaswani2017AttentionIA}. 
\begin{align}
    \mathbf{H}^{0}_{\mathbf{x}} &=  \texttt{one\_hot}(\mathbf{x})\cdot \mathbf{W}_e + \mathbf{W}_p \label{equ:token}\\
    \mathbf{H}^{l}_{\mathbf{x}} &= \texttt{trm\_block}(\mathbf{H}^{l-1}_{\mathbf{x}}), \ l=1,\dots, L
\end{align}
where 
$\mathbf{W}_e$ is the token embedding matrix, $\mathbf{W}_p$ is the position embedding matrix, $\texttt{trm\_block}(\cdot)$ is the transformer block with bi-directional attention and $L$ is the number of Transformer blocks. We typically choose the output of the last layer $\mathbf{H}^{L}_{\mathbf{x}}$ as the statement representation $\mathbf{H}_{\mathbf{x}}$.

Then we consider the following embeddings:
\begin{itemize}
    \item \textbf{Context-aware token embedding}. In order to enhance the contextual dependency of the concept $c$ to the statement $\mathbf{x}$, we utilize a bi-attention network~\cite{Seo2016BidirectionalAF} that models the cross interaction between the concept and the statement. 
    \begin{align}
        \mathbf{H}_c^{tok} &=  \texttt{one\_hot}(c)\cdot \mathbf{W}_e \\
        \mathbf{H}_c^{con} &= \texttt{bi-attention}(\mathbf{H}_c^{tok}, \mathbf{H}_\mathbf{x})
    \end{align}
    Then we integrate $\mathbf{H}_c^{tok}$ and $\mathbf{H}_c^{con}$ by max pooling and linear transformation to obtain a fixed-length representation that encodes the textual information of the concept $c$:
    \begin{equation}
        \mathbf{h}^{text}_c = \texttt{mlp}\Big(\texttt{max}\big( [\mathbf{H}^{tok}_c;\mathbf{H}^{con}_c]\big)\Big)
    \end{equation}
        \item \textbf{Concept distance embedding}. To encode the graph-aware structure information into the node representation, we design a concept distance embedding $\mathbf{h}^{dist}_c\in \mathbb{R}^{d_1}$ that encodes the relative distance from concept $c$ to the source concepts $\mathcal{C}_\mathbf{x}$ on the subgraph. 
    Specifically, the concept distance for concept $c$ is defined as the minimum length of the path that can be reached from one source concept in $\mathcal{C}_\mathbf{x}$:
    \begin{align}
        d_{c} &= \min_{c_\mathbf{x}\in \mathcal{C}_\mathbf{x}} \text{Dist}(c_\mathbf{x}, c) 
    \end{align}
    The concept distance is then used as an index to look up a trainable matrix $\mathbf{W}_d$ and obtain the $\mathbf{h}^{dist}_c\in \mathbb{R}^{d_1}$.
\end{itemize}

Finally, the fused concept representation $\mathbf{h}_c$ is obtained by concatenating the context-aware token embedding and the concept distance embedding.
\begin{equation}\label{equ:concept}
    \mathbf{h}_{c} = [\mathbf{h}^{text}_c; \mathbf{h}^{dist}_c]
\end{equation}
    
\subsection{Bridge Concept Extraction}\label{sec:concept-extraction}

We describe the core component of our method in this section, which extracts the bridge concepts for further explanation generation. 
It first scores triples on the subgraph to downweight the noisy paths. Then it aggregates the path scores to each connected concepts by a path routing process and deactivates the nodes with low routing scores. Finally it selects top-ranked bridge concepts from the activated nodes.

\subsubsection{Triple Scoring}

Firstly, we calculate the triple scores according to the representation of triples and the input statement. For each triple $e=(c_{e,head}, r_e, c_{e,tail})$ where $c_{e,head}$/$c_{e,tail}$ indicates the head / tail concept and $r_e$ denotes the relation, we can obtain its representation by concatenating the representations of the head concept, the relation and the tail concept:
\begin{equation}
    \mathbf{h}_{e} = [\mathbf{h}_{c_{e,head}}; \mathbf{h}_{r_e}; \mathbf{h}_{c_{e,tail}}] 
\end{equation}

Both the head and the tail representations are calculated by Equation (\ref{equ:concept}) and the relation representation is acquired by indexing a trainable relation embedding matrix $\mathbf{W}_r$. Then we use the statement representation to query each triple representation by taking the bilinear dot-product attention and calculate the selection probability for each triple:
\begin{align}
    \mathbf{h}_\mathbf{x} &= \texttt{max-pooling}(\mathbf{H}_\mathbf{x})\in \mathbb{R}^{d_1}\\
    P(\mathbf{e}|\mathbf{x}) &= \sigma(\mathbf{h}_{\mathbf{e}} \mathbf{W}_2 \mathbf{h}_\mathbf{x}^T)
\end{align}

We adopt weak supervision to supervise the triple scoring process. For each concept $c \in \mathcal{B}_{\mathbf{x}\rightarrow\mathbf{y}}$, 
we obtain the set of the shortest paths $\mathbf{P}_{\mathbf{x}\rightarrow c}$ using the breadth-first search from each concept of $\mathcal{C}_\mathbf{x}$ to $c$. We consider all these shortest paths $\mathbf{P}_{\mathbf{x}\rightarrow\mathbf{y}}=\bigcup_{c \in \mathcal{B}_{\mathbf{x}\rightarrow\mathbf{y}}} \mathbf{P}_{\mathbf{x}\rightarrow c}$ as the supervision of our triple scoring process as they connect the reasoning chain from the statement to the explanation with minimum distractive information. Accordingly, other triples in $G_{\mathbf{x}}$ which don't belong to $\mathbf{P}_{\mathbf{x}\rightarrow\mathbf{y}}$ are regarded as negative samples. The loss function of triple scoring is devised as follows:
\begin{align} \label{eqn:triple}
    \mathcal{L}_{triple} = &- \sum_{\mathbf{e}\in G_{\mathbf{x}}} \mathbb{I}(\mathbf{e}\in \mathbf{P}_{\mathbf{x}\rightarrow\mathbf{y}}) \log P(\mathbf{e}|\mathbf{x})\nonumber\\
    &+ [1-\mathbb{I}(\mathbf{e}\in \mathbf{P}_{\mathbf{x}\rightarrow\mathbf{y}})] \log [1-P(\mathbf{e}|\mathbf{x})]
\end{align}
where $\mathbb{I}(\mathbf{e}\in \mathbf{P}_{\mathbf{x}\rightarrow\mathbf{y}})$ is an indicator function that takes the value 1 iff $\mathbf{e}\in \mathbf{P}_{\mathbf{x}\rightarrow\mathbf{y}}$, and 0 otherwise.

\subsubsection{Path Routing}

Next, we describe the path routing process which involves propagating the scores along the paths to each concept on the subgraph from the source concepts. For each path $\mathbf{p}$ retrieved from the subgraph $G_\mathbf{x}$, we calculate a path score $s(\mathbf{p})$ by aggregating the triple score $P(\mathbf{e}|\mathbf{x})$ along the path:
\begin{equation}
    s(\mathbf{p}) = \frac{1}{|\mathbf{p}|}\sum_{\mathbf{e}\in \mathbf{p}} P(\mathbf{e}|\mathbf{x})
\end{equation}

For each concept $c$, we consider all the shortest paths $\mathbf{P}_{\mathbf{x}\rightarrow c}$ that starts with the source concepts and ends with $c$ monotonically, i.e., the concept distance of each node on the path increases monotonically along the path. Then we calculate the routing score for the concept $c$ by averaging the path scores of $\mathbf{P}_{\mathbf{x}\rightarrow c}$.
\begin{equation}
    s(c) = \frac{1}{|\mathbf{P}_{\mathbf{x}\rightarrow c}|}\sum_{\mathbf{p}\in \mathbf{P}_{\mathbf{x}\rightarrow c}} s(\mathbf{p})
\end{equation}

Intuitively, this process disseminates the triple scores and aggregates them to the connected concepts. 
Then we deactivate some paths based on the path routing results and obtain $V_{\mathbf{x}\rightarrow\mathbf{y}}$ by preserving concepts with the top-$K_1$ routing scores.

\subsubsection{Concept Selection}

Finally, we conduct concept selection based on the concept representation and the statement representation. For each concept in $V_{\mathbf{x}\rightarrow\mathbf{y}}$, we calculate the selection probability for it by taking the dot-product attention and adopt a similar cross-entropy loss with supervision from bridge concepts $\mathcal{B}_{\mathbf{x}\rightarrow\mathbf{y}}$:

\begin{align}
    P(c|\mathbf{x}) &= \sigma(\mathbf{h}_c \mathbf{W}_3 \mathbf{h}_\mathbf{x}^T) \\
    \mathcal{L}_{concept} &= - \sum_{\mathbf{c}\in V_{\mathbf{x}\rightarrow\mathbf{y}}} \mathbb{I}(\mathbf{c}\in \mathcal{B}_{\mathbf{x}\rightarrow\mathbf{y}}) \log P(c|\mathbf{x}) \nonumber\\
    &+ [1-\mathbb{I}(\mathbf{c}\in \mathcal{B}_{\mathbf{x}\rightarrow\mathbf{y}})]\log [1-P(c|\mathbf{x})]
\end{align}
where the indicator function is similar to that of Equation (\ref{eqn:triple}).

Finally, the bridge concepts with top-$K_2$ probability $P(c|\mathbf{x})$ are selected as the additional input to the generation model.

\subsection{Explanation Generation}\label{sec:generation}

We utilize a pre-trained Transformer decoder~\cite{Radford2019LanguageMA} as our generation model which shares the parameter with the Transformer encoder. Essentially, it takes the statement $\mathbf{x}$ and the concepts $\mathbf{c}$ as input and auto-regressively generates the explanation $\mathbf{y}$:
\begin{align}
    P(\mathbf{y}|\mathbf{x}, \mathbf{c}) &= P(\mathbf{y}|\mathbf{x}, c_1,\cdots, c_{K_2}) \nonumber\\ &= \prod_{t=1}^M P(\mathbf{y}_t|\mathbf{x}, c_1,\cdots, c_{K_2}, \mathbf{y}_{<t}) \\
    \mathcal{L}_{generation} &= -\log P(\mathbf{y}|\mathbf{x}, c_1,\cdots, c_{K_2})
\end{align}

As shown in Figure \ref{fig:model}, the input to the Transformer decoder is the token embeddings of both the statement and the selected concepts concatenated along the sequence length dimension.

To model bi-directional attention on the input side while preserving the causal dependency of the generated sequence, we adopt a hybrid attention mask where each token on the input side could attend to all the tokens in the input sequence while the generated token at each time step only attends to the input sequence and the previously generated tokens.

\subsection{Training and Inference}
To train the model, we optimize the final loss function which is the sum of the three loss functions:
\begin{equation}
    \mathcal{L}_{final} = \mathcal{L}_{generation} + \lambda_1 \mathcal{L}_{triple} + \lambda_2 \mathcal{L}_{concept}
\end{equation}

As for the inference process, Figure \ref{fig:model} demonstrates how our model retrieves reasoning paths given the statement, extracts bridge concepts and finally generates the explanation.

\section{Experiment}

\subsection{Dataset and Experimental Setup}

\subsubsection{Commonsense Explanation Dataset}

We adopt the dataset from the Commonsense Validation and Explanation Challenge\footnote{https://competitions.codalab.org/competitions/21080} which consists of three subtasks, i.e., commonsense validation, commonsense explanation selection and commonsense explanation generation. We focus on the explanation generation subtask in this paper. The commonsense explanation generation subtask contains $10,000$ statements that are against commonsense. For each statement, three human-written explanations are provided. To evaluate our proposed model and other baselines, we randomly split 10\% data as the test set, 5\% as the development set and the latter as the training set. Note that we further split each example in the training set into three statement-explanation pairs, while for the development set and the test set we use the three corresponding explanations as references for each statement. This results in our final data split (25,596 / 476 / 992) denoted as (train / dev / test). 

\subsubsection{Commonsense Knowledge Graph}

We use the English version ConceptNet as our external commonsense knowledge graph. It contains triples in the form of $(h, r, t)$ where $h$ and $t$ represent head and tail concepts and $r$ is the relation type. We follow \citet{Lin2019KagNetKG} to merge the original 42 relation types into 17 types. We additionally define 17 reverse types corresponding to the original 17 relation types to distinguish the direction of the triples on the graph.

\subsection{Automatic Evaluation Metrics}

To automatically evaluate the performance of the generation models, we use the \textbf{BLEU-3/4}~\cite{Papineni2001BleuAM}, \textbf{ROUGE-2/L}~\cite{Lin2004ROUGEAP}, \textbf{METEOR}~\cite{Banerjee2005METEORAA} as our main metrics. We also propose \textbf{Concept F1} to evaluate the accuracy of the unique concepts in the generated explanation that do not occur in the statement.

Specifically, given the generated explanation $\hat{\mathbf{y}}$ and the reference explanation $\mathbf{y}$, we extract a set of concepts $\mathcal{C}_{\hat{\mathbf{y}}}$ and $\mathcal{C}_{{\mathbf{y}}}$ from the generated explanation and the reference explanation respectively using the method in \S{\ref{sec:path-grounding}}. We denote the sets of unique concepts in the explanation as $\mathcal{U}_{{\mathbf{y}}}=\mathcal{C}_{{\mathbf{y}}}- \mathcal{C}_{{\mathbf{x}}}$ and $\mathcal{U}_{\hat{\mathbf{y}}}=\mathcal{C}_{\hat{\mathbf{y}}}- \mathcal{C}_{{\mathbf{x}}}$. Then we can compute the Concept F1 as the harmonic mean of \textit{recall} and \textit{precision}.
\begin{equation}
recall = \frac{|\mathcal{U}_{\hat{\mathbf{y}}} \cap \mathcal{U}_{\mathbf{y}}|}{|\mathcal{U}_{\mathbf{y}}|}, \ precision =  \frac{|\mathcal{U}_{\hat{\mathbf{y}}} \cap \mathcal{U}_{\mathbf{y}}|}{|\mathcal{U}_{\hat{\mathbf{y}}}|}
\end{equation}



\subsection{Implementation Details}

For the reasoning path retrieval process, we set the
maximum number of neighbours $B=300$ at each hop. For each example, we restrict the concepts of the subgraph to those only appeared in the training and development set. 

We use a pre-trained Transformer language model GPT-2~\cite{Radford2019LanguageMA} as the initialization of the Transformer model. We set the hidden dimension $d_1=768$ identical to the hidden size of the Transformer. We empirically set the following hyperparameters by tuning the model on the development set: selection threshold $K_1=30, K_2=3$, loss coefficients $\lambda_1=1, \lambda_2=1$, number of epochs $=3$, batch size $=4$, learning rate $=4\times 10^{-5}$ and use the Adam optimizer~\cite{kingma2014adam} with 10\% warmup steps. We select the model with the highest BLEU-4 score on the development set and evaluate it on the test set. At the decoding phase, we use beam search with a beam size of $3$ for all models.

\subsection{Baseline Models}

We compare with the following baseline models:
\begin{itemize}
    \item \textbf{Seq2Seq}: a sequence-to-sequence model based on gated recurrent unit (GRU) \cite{cho14gru} and attention mechanism, which is widely used in text generation tasks~\cite{bahdanau15attention}.
    \item \textbf{MemNet}: a knowledge-grounded sequence-to-sequence model~\cite{ghazvininejad18memnet}. In our experimental setting, we regard all the concepts which are connected with those in the statements as knowledge facts.
    \item \textbf{Transformer}: an encoder-decoder framework commonly used in machine translation tasks~\cite{Vaswani2017AttentionIA}.
    \item \textbf{GPT-2}: a multi-layer Transformer decoder pre-trained on WebText~\cite{Radford2019LanguageMA} which is then directly fine-tuned on our dataset.
\end{itemize}

\subsection{Experimental Results}

\begin{table}[t]
    \centering
    \small
    \begin{tabular}{ccccc}
    \toprule[1.5pt]
    \textbf{Model} & \textbf{B-3/4} & \textbf{R-2/L} & \textbf{M} & \textbf{Concept F1}\\
    \midrule[1pt]
    Seq2Seq & 10.7/6.1 & 9.9/25.8 & 11.4 & 11.1\\
    MemNet & 10.2/5.7 & 8.8/25.7 & 11.0 & 11.5\\
    Transformer & 10.0/5.8 & 9.6/26.0 & 12.0 & 11.7\\
    GPT-2-FT  & 23.4/15.7 & 18.9/36.5 & 17.7 & 17.4\\
    \midrule[1pt]
    Ours &  \textbf{24.7/17.1} & \textbf{20.2/37.9} & \textbf{18.3} & \textbf{20.1}\\
    \bottomrule[1.5pt]
    \end{tabular}
    \caption{Automatic evaluation of explanation generation in terms of BLEU (B), ROUGE (R), METEOR (M) and Concept F1.}
    \label{tab:auto-eval}
\end{table}

\begin{table}[t]
    \centering
    \small
    \begin{tabular}{lcc}
    \toprule[1.5pt]
    \textbf{Setting} & \textbf{BLEU-4} & \textbf{Concept F1}\\
    \midrule[1pt]
    Ours & 17.1 & 20.1 \\
    w/o Context Emb. & 16.0 & 18.6\\
    w/o Distance Emb. & 16.4 & 18.5\\
    w/o Path Routing & 16.5 & 19.2\\
    \midrule[1pt]
    \#Hop = 2 & 16.2 & 18.3\\
    \#Hop = 1 & 15.9 & 17.3\\
    \bottomrule[1.5pt]
    
    \end{tabular}
    \caption{Ablation study of our framework on the test set. 
    We present the model ablation results in the upper block and the data ablation results in the lower block.}
    \label{tab:ablation}
\end{table}

As shown in Table \ref{tab:auto-eval}, our model achieves the best performance in terms of all the automatic evaluation metrics, which demonstrates that our model can generate high quality explanations. Specifically, our model achieves a 2.7\% gain on Concept F1 compared with GPT-2 which indicates that explicitly extracting bridge concepts enhances the informativeness of the generated explanation. 

To evaluate the effects of different modules in our method, we conduct ablation studies on both the model components and the external knowledge base. For the model components, we test the following variants: (1) without the context-aware token embeddings (\textbf{w/o Context Emb.}); (2) without the concept distance embeddings (\textbf{w/o Distance Emb.}); (3) without the path routing process (\textbf{w/o Path Routing}). 
As for the data ablation, we sample subgraphs by restricting the maximum number of hops to 2 (\textbf{\#Hop=2}) and 1 (\textbf{\#Hop=1}).

As shown in Table \ref{tab:ablation}, each module contributes to the final results. Particularly, discarding the 
context-aware embeddings leads to the most remarkable performance drop, which indicates the significance for context modeling in multi-hop reasoning.
Besides, the data ablation results demonstrate that as the subgraph has less coverage, the generation model will suffer from the noisy concepts and thus deteriorate the generation results.

\begin{figure}
    \centering
    \includegraphics[width=0.8\columnwidth]{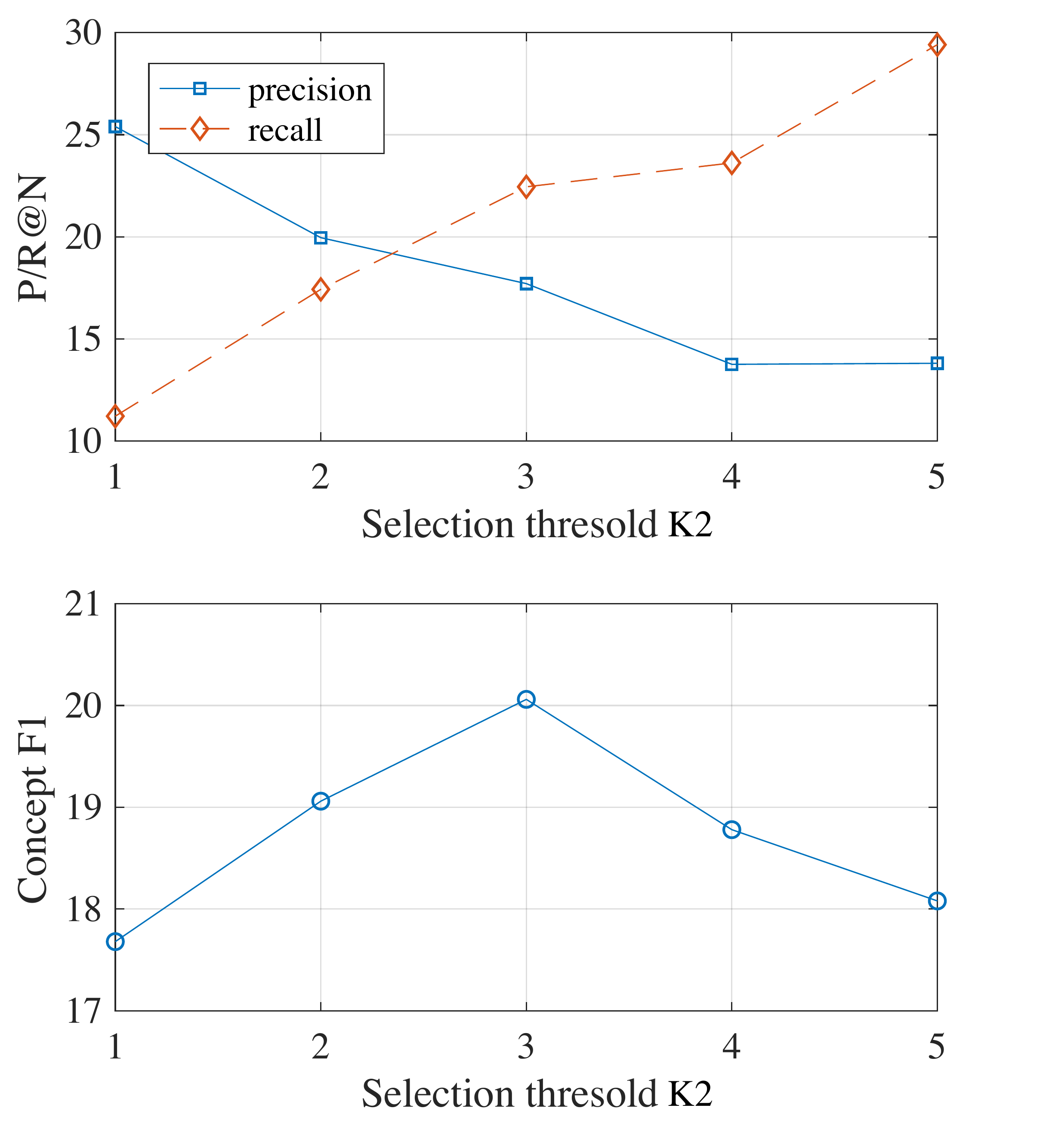}
    \caption{P/R@N measures the precision / recall of the top-$N$ selected bridge concepts. Concept F1 measures the F1-score of concepts in the generated explanations.}
    \label{fig:concept_stats}
\end{figure}

We additionally present the results of the selected and generated concepts with different concepts selection threshold $K_2$. As shown in the upper part of the Figure \ref{fig:concept_stats}, as the number of selected concepts increases, more true positives are selected, resulting in the increase of the recall (Recall@N) while the inclusion of more false positives leads to the decrease of the precision (Precision@N) . The Concept F1 reaches maximum when $K_2=3$ (see the lower part), which demonstrates that the model learns to extract critical concepts for explanation generation while keeping out most noisy candidates with an appropriate selection threshold.

\subsection{Human Evaluation}

\begin{table}[t]
    \centering
    \scriptsize
    \begin{tabular}{l  cc cc cc}
    \toprule[1.5pt]
        \multirow{2}*{\textbf{Model}} & \multicolumn{2}{c}{\textbf{Fluency}} & \multicolumn{2}{c}{\textbf{Reasonability}} & \multicolumn{2}{c}{\textbf{Informativeness}} \\
         & Win & Lose & Win & Lose & Win & Lose \\
         \midrule[1pt]
         vs. Seq2seq & 0.41 & 0.02 & 0.86 & 0.04 & 0.84 & 0.05 \\
        vs. MemNet & 0.48 & 0.00 & 0.84 & 0.03  & 0.87 & 0.03 \\
        vs. Transformer & 0.33 & 0.01  & 0.71 & 0.03 & 0.72 & 0.03 \\
        vs. GPT-2 & 0.20 & 0.10 & 0.40 & 0.27  & 0.34 & 0.15 \\
        \bottomrule[1.5pt]
    \end{tabular}
    \caption{Human evaluation results. The scores are the percentages of $win$ and $lose$ of our model in pair-wise comparison ($tie$ can be calculated by $1-win-lose$). Our model is significantly better (sign test, $\text{p-value} < 0.005$) than all the baseline models on all three criteria.}
    \label{tab:human}
\end{table}

\begin{table*}
\centering
    \small
    \begin{tabular}{llll}
    \toprule[1.5pt]
    \textbf{Error Type} & \textbf{Ratio (\%)} & \textbf{Input} & \textbf{Output}\\
    \midrule[1pt]
    \textbf{Repetition} & 7.7 & She begins working for relaxation. & People work to \underline{relax}, not \underline{relax}. \\
    \textbf{Overstatement} & 19.2 & Less people seek knowledge. & People \underline{don't} seek knowledge. \\
    \textbf{Unrelated} & 26.9 & The simplest carbohydrates are amino acid. & Alkaloids are not found in \underline{bread}. \\
    \textbf{Chaotic} & 11.5 & Giving assistance is for revenge. & If you help someone, you are \underline{grateful}. \\
    \bottomrule[1.5pt]
    \end{tabular}
    \caption{Distribution and typical cases of different error types of the explanations generated by our model. Underlined texts denote the error types including repetition, overstatement, unrelated words and chaotic expression.}
     \label{tab:error}
\end{table*}

\begin{table}[h]
    \centering
    \small
    \begin{tabular}{l}
        \toprule[1.5pt]
        \textbf{Statement 1:} I \textcolor{green!70!black}{buy} popcorn and knife when I go to the \\
        \textcolor{green!70!black}{cinema}.\\
        \midrule[1pt]
        \textbf{Seq2Seq:} A person cannot \textcolor{red!90!black}{buy a person} to watch a movie.\\
        \textbf{MemNet:} A \textcolor{red!90!black}{toothbrush} is not a place to play a movie.\\
        \textbf{Transformer:} A \textcolor{red!90!black}{fridge} is not a place to store \textcolor{red!90!black}{groceries}.\\
        \textbf{GPT-2:} You don't buy popcorn and knife at the cinema.\\
        \midrule[1pt]
        \textbf{Ours:} Knives are not \textcolor{green!70!black}{\textbf{sold}} at the cinema.\\
        \textbf{Top-3 reasoning paths:} (buy$\rightarrow$\textit{antonym}$\rightarrow$sell), (popcorn\\ $\rightarrow$ \textit{related to}$\rightarrow$food), (cinema$\rightarrow$ \textit{related to}$\rightarrow$movie)\\
        \textbf{Selected concepts:} \textcolor{green!70!black}{\textbf{sell}}, place, movie \\
        \midrule[1pt]
        \textbf{Statement 2:} He \textcolor{green!70!black}{eats} his chips with \textcolor{green!70!black}{toothpaste}. \\
        \midrule[1pt]
        \textbf{Seq2Seq:} \textcolor{red!90!black}{Chopsticks} are not edible. \\
        \textbf{MemNet:} A \textcolor{red!90!black}{potato} is too soft to eat \textcolor{red!90!black}{juice} with your teeth. \\
        \textbf{Transformer:} You do not eat \textcolor{red!90!black}{sand with a cup}.\\
        \textbf{GPT-2:} Toothpaste is not edible. \\
        \midrule[1pt]
        \textbf{Ours:} Toothpaste is \textcolor{green!70!black}{used} to clean \textcolor{green!70!black}{teeth}. \\
        \textbf{Top-3 reasoning paths:} (eat$\rightarrow$\textit{related to}$\rightarrow$tooth), (\\ toothpaste$\rightarrow$\textit{related to}$\rightarrow$paste$\rightarrow$\textit{related to}$\rightarrow$use), (eat\\ 
        $\rightarrow$\textit{has subevent} $\rightarrow$work$\rightarrow$\textit{related to}$\rightarrow$use) \\
        \textbf{Selected concepts:} \textcolor{green!70!black}{\textbf{use}}, \textcolor{green!70!black}{\textbf{tooth}}, food \\ 
        \bottomrule[1.5pt]
         
    \end{tabular}
    \caption{Examples of generated explanations. Irrelevant contents are in red and critical concepts for explanation are in green.}
    \label{tab:case}
\end{table}

To further evaluate the quality of the generated explanations, we conduct the human evaluation and recruit five annotators to perform pair-wise comparisons. Each annotator is given 100 paired explanations (one generated by our model and the other by a baseline model, along with the statement) and is required to give a preference among ``win'', ``tie'', and ``lose'' according to three criteria: (1) \textit{Fluency} which measures the grammatical correctness and the readability of the explanation. (2) \textit{Reasonability} which measures whether the explanation is reasonable and accords with the commonsense. (3) \textit{Informativeness} which measures the amount of new information delivered in the explanation that helps explain the statement.

The results are shown in Table \ref{tab:human}, our model outperforms all the baseline models significantly on all three criteria (sign test, $\text{p-value} < 0.005$). Specifically, our model wins GPT-2 substantially in terms of reasonability and informativeness.

To evaluate the inter-rater agreement for each criterion, we calculate the Fleiss' kappa~\cite{Fleiss1971MeasuringNS}. For \textit{Reasonability} / \textit{Informativeness}, the kappa is 0.429 / 0.433 respectively indicating a moderate agreement among annotators. In terms of \textit{Fluency}, annotators show diverse preferences ($\kappa=0.245$) since GPT-2 has strong ability in generating fluent texts.

\subsection{Case Study}

Table \ref{tab:case} presents the generated explanations. Our model is capable to generate reasonable and informative explanations by utilizing the extracted bridge concepts. Specifically, in the first case our model extracts bridge concepts ``sell'' and identifies the incompatibility between ``knives'' and ``cinema''. In the second case, our model clarifies the function of the ``toothpaste'' by extracting ``use'' from two reasoning paths and provides more information rather than simply negative phrasing.

\subsection{Error Analysis}

To analyze the error types of the explanations generated by our model, we manually check all the failed cases\footnote{The decision is based on majority voting by the five annotators.} in the pair-wise comparison between our model and the strong baseline GPT-2. The number of these cases is 26 in all 100 explanations. We manually annotated four types of errors from the failed explanations: \textbf{repetition} (words repeating), \textbf{overstatement} (overstate the points), \textbf{unrelated} concepts towards the statement (the explanation itself may be reasonable), \textbf{chaotic} sentences (difficult to understand). As shown in Table \ref{tab:error}, 
it is still challenging for the model to generate explanations highly related to the statement with accurate wording.

\section{Conclusion}

In this paper, we analyze the challenges in incorporating external knowledge graph to aid the commonsense generation problem and propose a two-stage method that first extracts bridge concepts from a retrieved subgraph and then generates the explanation by integrating the extracted concepts. Experimental results show that our model outperforms baselines including the strong pre-trained language model GPT-2 in both automatic and manual evaluation. 

\section*{Acknowledgments}

This work was jointly supported by the NSFC projects (key project with No. 61936010 and regular project with No. 61876096), and the Guoqiang Institute of Tsinghua University with Grant No. 2019GQG1. We thank THUNUS NExT Joint-Lab for the support.

\bibliography{aacl-ijcnlp2020}
\bibliographystyle{acl_natbib}

\end{document}